\documentclass[letterpaper, 10 pt, conference]{ieeeconf}  % Comment this line out if you need a4paper
\IEEEoverridecommandlockouts                              % This command is only
\usepackage{graphicx,epstopdf}

\usepackage{bm}
\usepackage{amssymb,amsthm}
\usepackage{amsmath}
\usepackage{amsfonts}
\usepackage{mathrsfs}
\usepackage{tabularx} 
\usepackage[linesnumbered,boxed,ruled,commentsnumbered]{algorithm2e}

\usepackage{acronym}
\usepackage{todonotes}
\usepackage{threeparttable}
\usepackage{color}
\usepackage{mathtools}
\usepackage{booktabs}
\usepackage{multirow}
\usepackage[caption=false,font=footnotesize]{subfig}
\newif\ifuseboldmathops
\newif\ifuseittextabbrevs
\useboldmathopstrue   % comment out to use mathbb
\ifuseittextabbrevs
	
	\newcommand{\eg}{{\it e.g. }}
	\newcommand{\ie}{{\it i.e. }}

\else
	
	\newcommand{\eg}{e.g. }
	\newcommand{\ie}{i.e. }

\fi

\ifuseboldmathops
\else

\fi

\ifuseboldmathops

\else

\fi

\ifuseboldmathops
	\newcommand{\Expect}{\mathop{\mathbb{E}{}}\nolimits}
	
\else
	\newcommand{\Expect}{\mathop{\mathbb{E}{}}\nolimits}
	
\fi

\ifuseboldmathops

\else

\fi

\newcommand{\union}{\bigcup}

\newcommand{\calL}{\mathcal{L}}
\newcommand{\calF}{\mathcal{F}}
\newcommand{\calN}{\mathcal{N}}

\newcommand{\calA}{\mathcal{A}}
\newcommand{\calS}{\mathcal{S}}
\newcommand{\calG}{\mathcal{G}}
\newcommand{\calE}{\mathcal{E}}

\newcommand{\calP}{\mathcal{P}}

\newcommand{\calQ}{\mathcal{Q}}
\newcommand{\calD}{\mathcal{D}}

\newcommand{\iactor}[1]{\pi^{\boldtheta_{#1}}}
\newcommand{\icritic}[1]{\calQ^{\boldphi_{#1}}}
\newcommand{\icost}[1]{\calQ^{\boldomega_{#1}}}

\newcommand{\JR}[2]{J^R_{#1}(#2)}
\newcommand{\JC}[2]{J^C_{#1}(#2)}
\newcommand{\dotprod}[2]{\langle #1, #2 \rangle}
\newcommand{\hatgrad}{\widehat{\grad}}

\newcommand{\avg}[2]{\text{Avg}_{#1}(#2)}

\newcommand{\grad}{\nabla}
\newcommand{\boldtheta}{\boldsymbol{\theta}}
\newcommand{\boldvartheta}{\boldsymbol{\vartheta}}
\newcommand{\boldlambda}{\boldsymbol{\lambda}}
\newcommand{\boldphi}{\boldsymbol{\phi}}
\newcommand{\boldomega}{\boldsymbol{\omega}}

\renewcommand{\vec}[1]{\bm{#1}}

\newcommand{\safe}{\text{safe}}
\newcommand{\INF}{\text{inf}}
\renewcommand{\det}{\text{det}}

\newcommand{\KLS}{\text{KEEP-LANE-SPEED}}
\newcommand{\CLL}{\text{CHANGE-LANE-LEFT}}
\newcommand{\CLR}{\text{CHANGE-LANE-RIGHT}}

\acrodef{mdp}[MDP]{Markov decision process}
\acrodef{dfa}[DFA]{deterministic finite-state automaton}
\acrodef{ltl}[LTL]{linear temporal logic}
\acrodef{ltlf}[LTL$(\calF)$]{quantitative linear temporal logic}
\acrodef{ag}[AG]{Assume-Guarantee}
\acrodef{ssp}[SSP]{Stochastic Shortest Path}
\acrodef{mcmc}[mcmc]{Monte Carlo Markov chain}

\theoremstyle{definition}
 \newtheorem{definition}{Definition}
 \newtheorem{example}{Example}

\acrodef{ltl}[LTL]{linear temporal logic formula}
\acrodef{mdp}[MDP]{Markov decision process}
\acrodef{smdp}[Semi-MDP]{Semi-Markov decision process}

\acrodef{scltl}[scLTL]{syntactically co-safe LTL}
\begin{document}
\title{\LARGE \bf Spatial-Temporal-Aware Safe Multi-Agent Reinforcement Learning of Connected Autonomous Vehicles in Challenging Scenarios}
\author{Zhili Zhang \and Songyang Han \and Jiangwei Wang \and Fei Miao
\thanks{This work was supported by NSF 1849246, NSF 1932250, NSF 2047354 grants. Z.~Zhang, S.~Han, and F.~Miao are with the Department of Computer Science and Engineering, J.~Wang is with the Department of Electrical and Computer Engineering, University of Connecticut, Storrs Mansfield, CT, USA 06268.  Email: \{zhili.zhang, songyang.han, jiangwei.wang, fei.miao\}@uconn.edu.}}
%\date{}							% Activate to display a given date or no date

\maketitle

\begin{abstract}
Communication technologies enable coordination among connected and autonomous vehicles (CAVs). However, it remains unclear how to utilize shared information to improve the safety and efficiency of the CAV system. In this work, we propose a framework of constrained multi-agent reinforcement learning (MARL) with a parallel safety shield for CAVs in challenging driving scenarios. The coordination mechanisms of the proposed MARL include information sharing and cooperative policy learning, with Graph Convolutional Network (GCN)-Transformer as a spatial-temporal encoder that enhances the agent's environment awareness. The safety shield module with Control Barrier Functions (CBF)-based safety checking protects the agents from taking unsafe actions. We design a constrained multi-agent advantage actor-critic (CMAA2C) algorithm to train safe and cooperative policies for CAVs. With the experiment deployed in the CARLA simulator, we verify the effectiveness of the safety checking, spatial-temporal encoder, and coordination mechanisms designed in our method by comparative experiments in several challenging scenarios with the defined hazard vehicles (HAZV). Results show that our proposed methodology significantly increases system safety and efficiency in challenging scenarios.
\end{abstract}
\section{Introduction}
\label{sec:intro}
Wireless communication technologies such as WiFi and 5G cellular networks enable vehicle-to-everything (V2X) communication and help the autonomous vehicle to get extra information about the driving environment beyond its sensing capability~\cite{martin2020low,mun2021secure}. Shared information captured by the onboard sensors such as cameras and LIDARs-based vision information can be used to improve connected autonomous vehicles' (CAVs) decision-making~\cite {buckman2020generating,miller2020cooperative, han2022stable}. Shared basic safety messages (BSMs) (velocity, position, heading angle, and yaw rate) benefit the coordination and control decisions of CAVs in scenarios such as cross intersections and lane-merging~\cite{Coordinate_CAV, CV_intersection}.  

% discusses how a vehicle with high-fidelity sensors can help a vehicle with low-fidelity sensors for better perception and localization by shared vision, but without considering how to use shared information for behavior planning. The work~\cite{kim2015impact} shows how shared vision can provide a see-through or satellite view to human drivers based on experiments for reactive early lane changing, however, they do not analyze how autonomous vehicles can exploit the see-through view. 
%Work in \cite{PredictPolicyLearn_LeCun} uses vision to predict trajectory and speed for reinforcement learning training, but it seems from the video the cars do not change lane. In our work, we consider how CAVs can utilize shared information including vision on a multi-lane road. The work in \cite{chen2015deepdriving} maps from an image to affordance, then apply a simple car-following rule.

\begin{figure}[!t]
\centering
\subfloat[]{\centering\label{fig:its_init}\includegraphics[height=2.8cm,width=2.8cm]{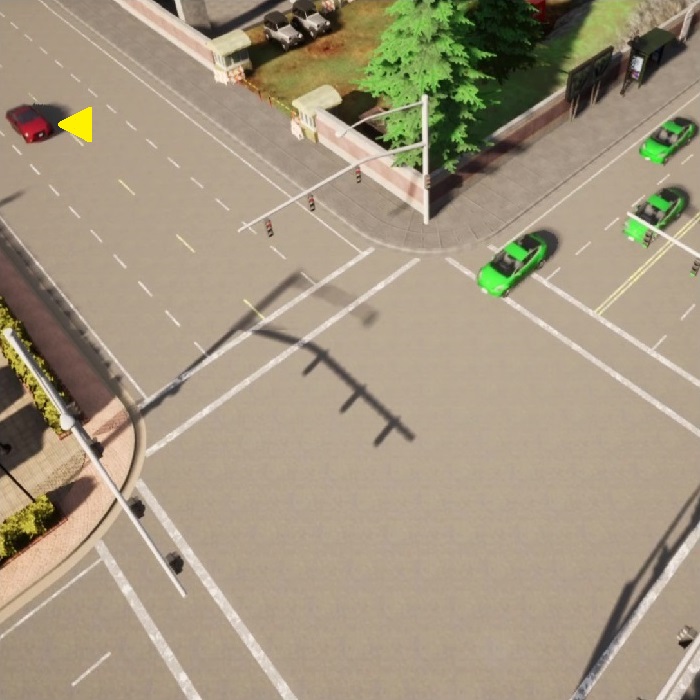}}\hfill
\subfloat[]{\centering\label{fig:its_succ}\includegraphics[height=2.8cm,width=2.8cm]{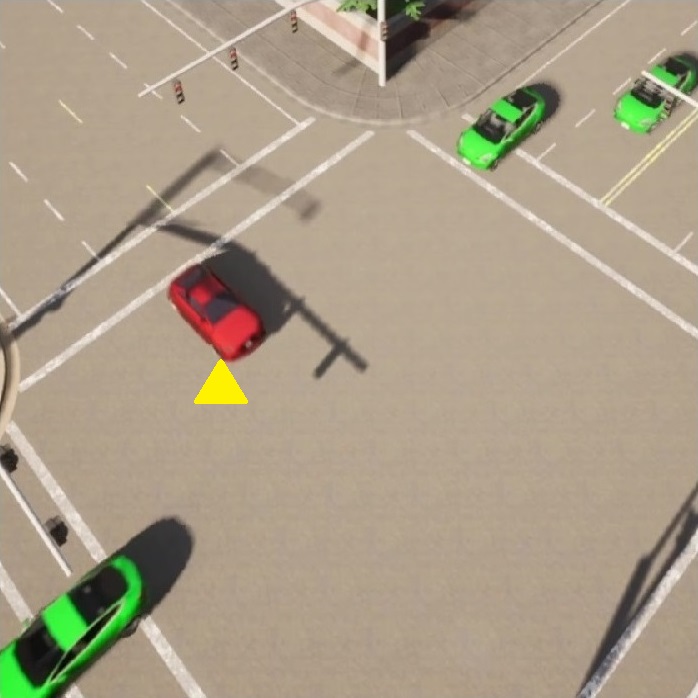}}\hfill
\subfloat[]{\centering\label{fig:its_fail}\includegraphics[height=2.8cm,width=2.8cm]{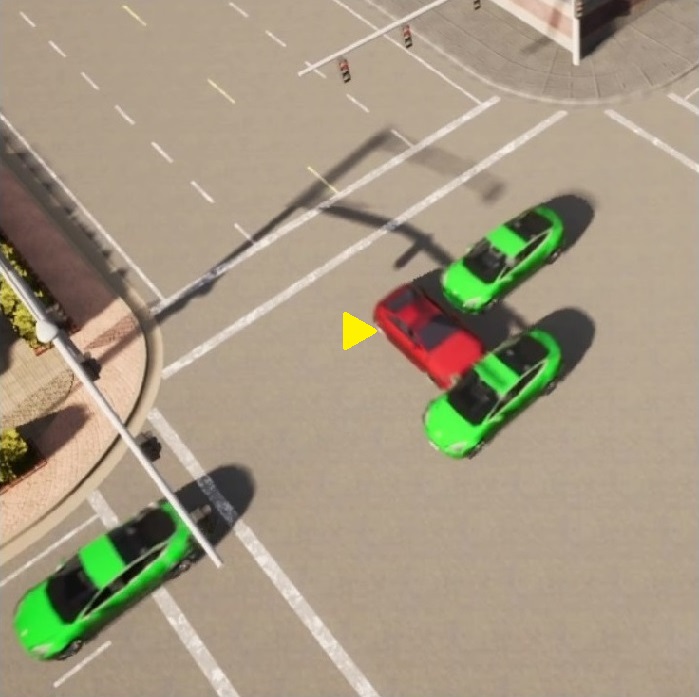}}
\\
\vspace*{-0.2cm}
\subfloat[]{\centering\label{fig:hwy_init}\includegraphics[height=2.8cm,width=2.8cm]{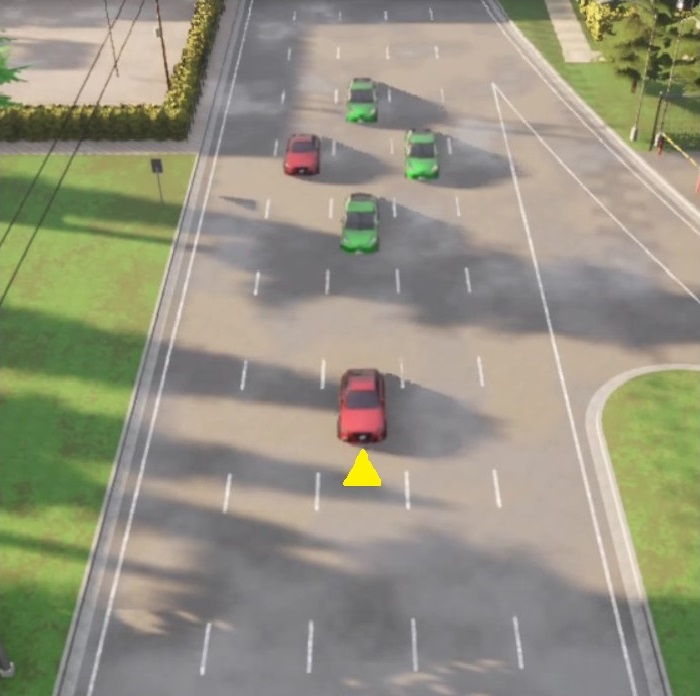}}\hfill
\subfloat[]{\centering\label{fig:hwy_succ}\includegraphics[height=2.8cm,width=2.8cm]{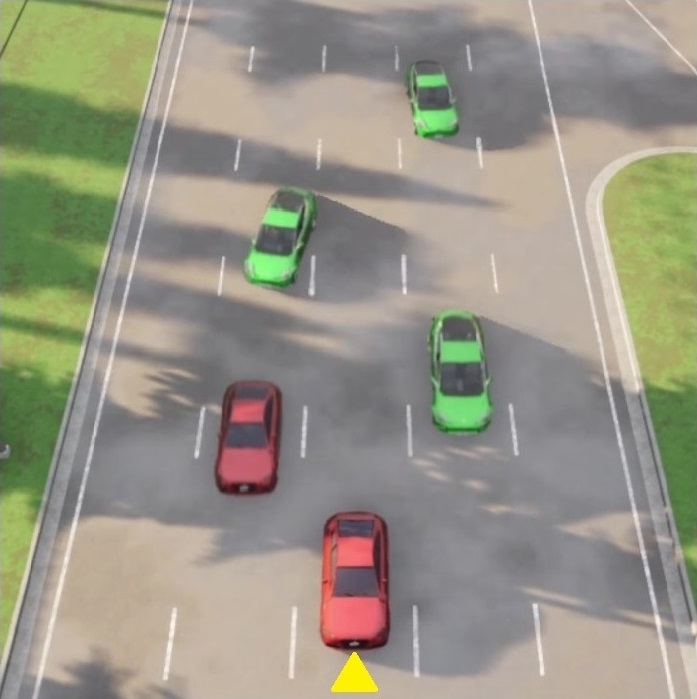}}\hfill
\subfloat[]{\centering\label{fig:hwy_fail}\includegraphics[height=2.8cm,width=2.8cm]{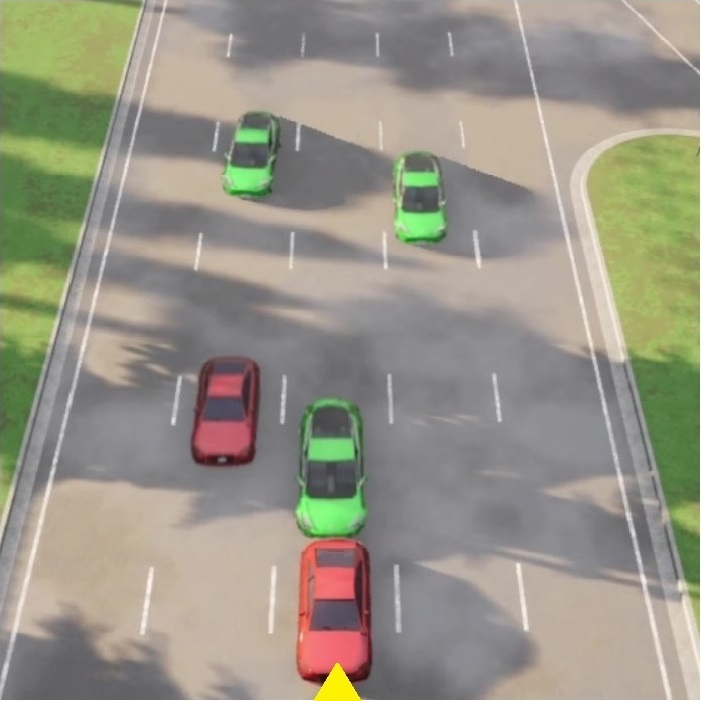}}
%\vspace*{-0.2cm}
\caption{\textit{Intersection} (upper) and \textit{Highway} (lower) scenarios. ~\ref{fig:its_init},~\ref{fig:hwy_init}: scenario initialization; ~\ref{fig:its_succ},~\ref{fig:hwy_succ}: successful cases of collaborative collision-avoidance from test runs of our method; ~\ref{fig:its_fail},~\ref{fig:hwy_fail}: collision cases from test runs of baseline model. Connected autonomous vehicles (CAVs) are in green; unconnected vehicles (UCVs) are in red; the hazard vehicle (HAZV) is in red with a yellow triangle mark. The hazard vehicle runs the red light in \textit{Intersection} scenario and takes a sudden hard-brake in \textit{Highway}. Without the safety shield or coordination, CAVs are likely to collide with HAZV or other vehicles as in~\ref{fig:its_fail},~\ref{fig:hwy_fail}.}
\label{fig:Scenarios}
\vspace{-0.5cm}
\end{figure}

However, it is not clear how information sharing benefits connected autonomous vehicles in challenging scenarios. Without communication and coordination, it is difficult for CAVs to react to a traffic-rule-violating behavior or sudden acceleration/deceleration maneuvers taken by the hazard vehicle as shown in Fig.~\ref{fig:Scenarios}. When an autonomous vehicle gets extra knowledge about the environment via coordinated V2X communication, how to design the neural network structure to utilize the shared information with spatial and temporal features and how to make prudent decisions to improve collaborative safety are unsolved challenges. %Hence, we show how CAVs can take advantage of information sharing to make better driving decisions while meeting safety guarantees and improving traffic speed in CARLA (Car Learning to Act)~\cite{Dosovitskiy17}, an open-source simulator for autonomous vehicle research. Illuminated by existing works of reinforcement learning with safety constraints \cite{10.5555/3305381.3305384, wen2020safe, lu2021decentralized}, 

In this work, we design a spatial-temporal-aware constrained MARL framework with parallel \textit{Safety Shield} for cooperative policy-learning of CAVs, to improve the safety and efficiency of the system utilizing V2X communication-based information-sharing. In particular, we consider challenging driving scenarios with potential traffic hazard vehicles. The complicated dynamics and interactions among CAVs under challenging scenarios provide strong motivation for us to design a \textit{Safety Shield} for the actions and policies of MARL, introduced in~\ref{sec:Safety_Checking}. We further introduce coordination mechanisms, as illustrated in Fig~\ref{fig:framework}. We utilize the prevailing Graph Convolutional Network (GCN) and Transformer structures as spatial-temporal scene encoders (Fig.~\ref{fig:pipeline}) for each agent to raise their situation awareness, as the actor-critic-cost neural network of the MARL model.
%(e.g. vehicles that violate traffic rules such as driving across the intersection during red light, have very large acceleration compared with other vehicles)
In summary, the main contributions of this work are:
\begin{itemize}
\item We propose a framework of constrained MARL with the designed \textit{Safety Shield} based on Control Barrier Functions (CBFs) and verify the significant improvement in collision-free rate with experiments.
\item We design a GCN-Transformer encoder integrated with MARL to utilize the shared spatial and temporal information among CAVs. Compared with the baseline model, our solution is enabled to achieve higher safety metrics and overall returns in challenging scenarios.
 %Applying GCN+Transformer as the spatial-temporal encoder supporting MARL decisions. 
\item We introduce coordination mechanisms to MARL with information-sharing and cooperative policy-learning. Our experiment results show that cooperation among CAVs improves the collision-free rate and overall return.
\end{itemize}

\section{Related Work}
\label{sec:related}
\paragraph{Planning and Control of Autonomous Vehicles} To learn the output control signals for steering angle and acceleration directly based on the observed environment, end-to-end learning is designed in CNN-based supervised learning~\cite{bojarski2016end}, and CBF-based Deep Reinforcement Learning~\cite{cheng2019end}, when only considering lane-keeping without lane-changing behavior. 
%Model predictive control (MPC) has been used in supervised learning~\cite{chen2018approximating} and guided policy search for the DRL~\cite{MPCGPS_icra16}. 
The other popular way is to separate the learning and control phases. Learning methods can give a high-level decision, such as ``go straight", ``go left"~\cite{pan2017virtual}, or whether or not to yield to another vehicle~\cite{shalev2016safe}. It also works to first extract image features and then apply control upon these features~\cite{chen2015deepdriving}. However, the works mentioned above do not consider the connection between CAVs, while we consider how CAVs should use information sharing to improve the safety and efficiency of the system, and design an MARL-based algorithm such that CAVs cooperatively take actions under challenging driving scenarios.

\paragraph{GCN, Transformer and Deep MARL}  It has not been addressed yet how to specifically design a neural network structure to utilize the communication among CAVs to improve the system's safety or efficiency in policy learning. Recent advances like GCN~\cite{malawade2022spatiotemporal} and Transformer~\cite{iqbal2019actor,zhao2021spatial} show their advantages in processing spatial and temporal properties of data. We utilize a GCN-Transformer structure to capture the spatial-temporal information of driving scenarios to improve the coordination among CAVs. To the best of our knowledge, we are the first to design a GCN-Transformer structure-based deep constrained MARL framework to utilize the shared information among CAVs. We validate that this design improves the safety rates and total rewards for CAVs in challenging scenarios with traffic hazards.

\paragraph{Constrained MDP and Safe RL} Existing multi-agent reinforcement learning (MARL) literature~\cite{zhang2019multi,foerster2017counterfactual, rashid2018qmix, lowe2017multi} has not fully solved the challenges for CAVs. Constrained Markov Decision Process (CMDP)~\cite{wen2020safe, lu2021decentralized} learns a policy to maximize the total reward while maintaining the total cost under certain constraints. However, the cost or the constraint does not explicitly represents all the safety requirements of physical dynamic systems and cannot be directly applied to solve CAV challenges. The recent advance with a formal safety guarantee is the model predictive shielding (MPS) that also works for multi-agent systems~\cite{li2020robust, zhang2019mamps}. However, their safety guarantee assumes an accurate model of vehicles which is difficult to find in reality. Control Barrier Functions are used to map unsafe actions to a safe action set in MARL~\cite{han2022behavior}, but they do not consider how to design a spatial-temporal encoder actor or critic network structure for challenging scenarios with hazard vehicles. In this work, we first integrate the strengths of both constrained MARL and CBF-based safety shield to further improve the safety of CAVs under the threat of traffic hazards. %The coordination among CAVs in our proposed CMAA2C algorithm  %\textcolor{blue}{We propose a constrained MARL with safety shield to 
%\textcolor{red}{We also design a constrained actor-critic algorithm to....}

% We consider a challenging CAV problem with information sharing and design a constrained multi-agent actor-critic algorithm with  structure.

\section{Problem Formulation}
\label{sec:prob}
\subsection{Problem Description}
We consider the cooperative policy-learning problem for CAVs in challenging scenarios occurred on a multi-lane urban intersection or on a multi-lane highway (as shown in Fig.\ref{fig:Scenarios}). Other traffic participants include unconnected vehicle (UCVs) and a hazard vehicle (HAZV). Meanwhile infrastructures that have sensing, communication and computation abilities also play a supportive role to CAVs. 

A CAV agent is primarily supported with its own observation $o_i$, the shared observation $o_{\calN_i}$ from neighboring agents $\calN_i$ based on V2V communication and the shared observation $o_{\INF}$ from the road infrastructures. Specifically, $\calN_i$ provides extra sensor measurements and sensor-detection data, such as lane-detection with camera images and object detection with LiDARs~\cite{li2022v2x}. $o_{\INF}$ is broadcasted messages to CAVs from road infrastructures, such as Radar that can broadcast the detected speed and location of nearby vehicles.

\subsection{Constrained MARL Problem Formulation}
A Constrained MARL is defined as a tuple $G = (\calS, \calA, P, \{r_i\}, \{c_i\}, \calG, \gamma)$ where $\calG\coloneqq(\calN, \calE)$ is the communication network of all CAV agents;
% cost/constraint in previous work 
$\calS$ is the joint state space of all agents: $\calS:=\calS_1 \times \dots \times \calS_n$. The state space of agent $i$: $\calS_i = \{o_i, o_{j \in \mathcal{N}_i}, o_{\INF}\}$ contains information from three sources: self-observation $o_i$ from vehicle $i$'s own odometers and sensors, observation $o_{j \in \calN_i}$ shared by other connected agents and observation $o_{\INF}$ shared by infrastructure. The observation of each CAV is $o_i = \{(\vec{l}_i, \vec{v}_i, \vec{\alpha}_i), \det_{i} \}$, where $(\vec{l}_i, \vec{v}_i, \vec{\alpha}_i)$ is the GPS location, velocity and acceleration of agent $i$, $\det_i$ is the vision-based sensors (on-board camera and 3D point-cloud LiDAR) object detection results. %, such as lane index. 
The joint action set is $\calA:= \calA_1\times \cdots \times \calA_n$ where $\calA_i = \{a_{i,1}, a_{i,2}, \cdots, a_{i,4+k} \}$ is the discrete finite action space for agent $i$, and
\begin{itemize}
    \item $a_{i,1}$: $\KLS$ - the CAV $i$ maintains current speed in the current lane 
    \item $a_{i,2}$: $\CLL$ - the CAV $i$ changes to its left lane. In experiment, by taking $a_{i,2}$ we set the target waypoint on the left lane.
    \item $a_{i,3}$: $\CLR$ - the CAV $i$ changes to its right lane. In experiment, by taking $a_{i,3}$ we set the target waypoint on the right lane.
    \item $a_{i,4}$: BRAKE. In the experiment, the CAV $i$'s actuator will compute a brake value within range $brake^t_i\in [0, 0.5]$ at time $t$.
    \item $a_{i,5}, a_{i,6}, \ldots, a_{i,4+k}$ are $k$ discretized throttle intervals. Given the available throttle value set in the simulator as $[0,1]$, we set 
    $a_{i,4+j}=[\frac{j-1}{k},\frac{j}{k}]$. By choosing the action $a_{i,5}$, for example, the actuator of the vehicle $i$ will maintain in current lane and compute a throttle value $throttle_i \in [\frac{j-1}{k},\frac{j}{k}]$ according to controller's approach.
%$\sim \mathcal{N}(\mu_{a4},\,\sigma_{a4}^{2})$, $\mu_{a4} = \frac{0. + 0.3}{2}$, $\sigma_{a4}^{2} = \frac{0.3-0.}{2}$  
\end{itemize}
The state transition function is $P: \calS\times \calA \times \calS \mapsto [0,1]$.
The reward function $r\coloneqq \calS\times\calA \mapsto\mathbb{R}$. With agent $j$'s velocity defined as $\vec{v}_j$, agent $i$'s reward function is $r_i(s,a)=\sum_{j\in\calN}\mu_{i,j}\|\vec{v}_j\|_2$, with $\mu_{i,j}$ as non-negative weights. Every agent aims to maximize the weighted sum of all agents' speed. The cost function $c_i\coloneqq \calS\times\calA \mapsto\mathbb{R}$ is defined as $c_i(s,a) = \min(\|\vec{l}_i-\vec{l}_j\|,\|\vec{l}'_i-\vec{l}'_j\|\ |\ \forall j\in \{\calN_{-i}\cup O_i\})$, in which we consider the ego vehicle's distance to its closest neighbor and all the detected environment vehicles $O_i$ for the current step location $\vec{l}_i$ and next step location $\vec{l}'_i$. The local policy used by agent $i$ is defined as: $\iactor{i}(a_i|s_i)$. $\gamma=(\gamma_r, \gamma_c)$ are the discount factors for reward and cost respectively.

%\zhili{WRITE!!!!!!!!!}
\subsection{Spatial-Temporal Encoding}
Graph Convolutional Network and Transformer~\cite{iqbal2019actor,zhao2021spatial} have shown their advantages in modeling spatial and sequential information. GCN has been utilized~\cite{malawade2022spatiotemporal, schmidt2022crat, 9811567} to decode interactions between vehicles for collision and trajectory predictions. In this work, as shown in Fig.~\ref{fig:pipeline}, we design a GCN-Transformer module utilizing shared information to encode spatial-temporal features of driving environment, and feed each agent's reinforcement learning model. Ego vehicle's observation $o_i$, shared observations $o_j$'s and $o_{\inf}$ from V2X communication are used to construct graphs comprised of vehicles, roads and intersections, and edges among them. Vehicles are connected to their neighbors and their located road or intersection. With such graphs in consecutive time steps as input, the GCN-Transformer module encodes each agent $i$'s dynamic $(\vec{l}_i, \vec{v}_i, \vec{\alpha}_i)$ with their graph neighbors and generates the spatial-temporal representation of the environment as the input to the MARL model.
%In this paper, we use GCN and transformer directly for the decisions of driving behavior. Preceding works using Graph Networks in the field of autonomous vehicle usually focus on a specific task. 

%\subsubsection{Graph Convolutional Network}

%\subsubsection{Transformer}

% explain in lang
% explain in math formulation
% proof of reason

\section{Methodology}\label{sec:Methodology}
\begin{figure*}[!t]
\centering
\subfloat[{Model pipeline for a single agent. The state information as time series $\{s^{t-\tau}\}_{\tau}$ will be processed as graphs first and sequentially enter the GCN-Transformer module and the Actor's policy network; meanwhile, $s^t$ is input to the CBF safety checking module for computing safe actions. During training, the outputs of GCN-Tranformer will be input to the Critic and Cost network for advantage, constraint and TD error calculation.}]{\centering\includegraphics[width=12.5cm, height=4.8cm]{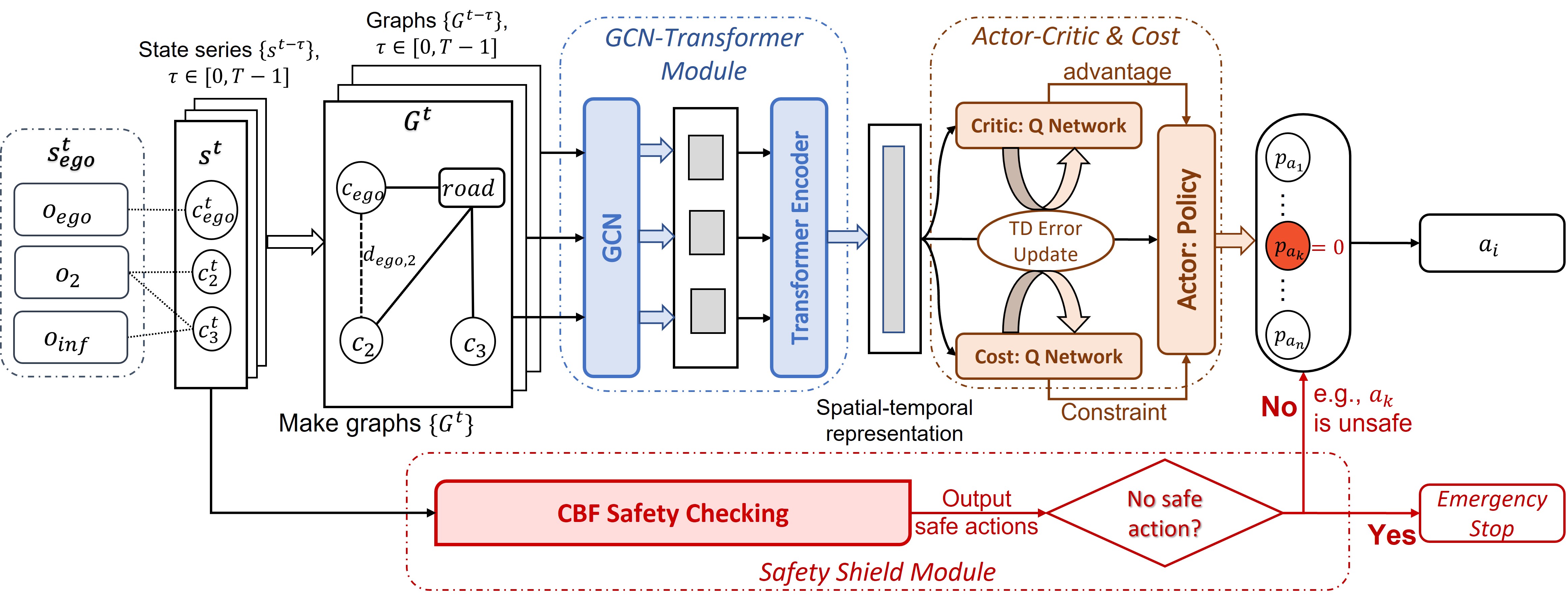}
\label{fig:pipeline}}
\hfill
\subfloat[{Constrained MARL and safety shield framework, with coordinated information-sharing and policy-learning among agents.}]{\centering\includegraphics[width=4.5cm, height=4.8cm]{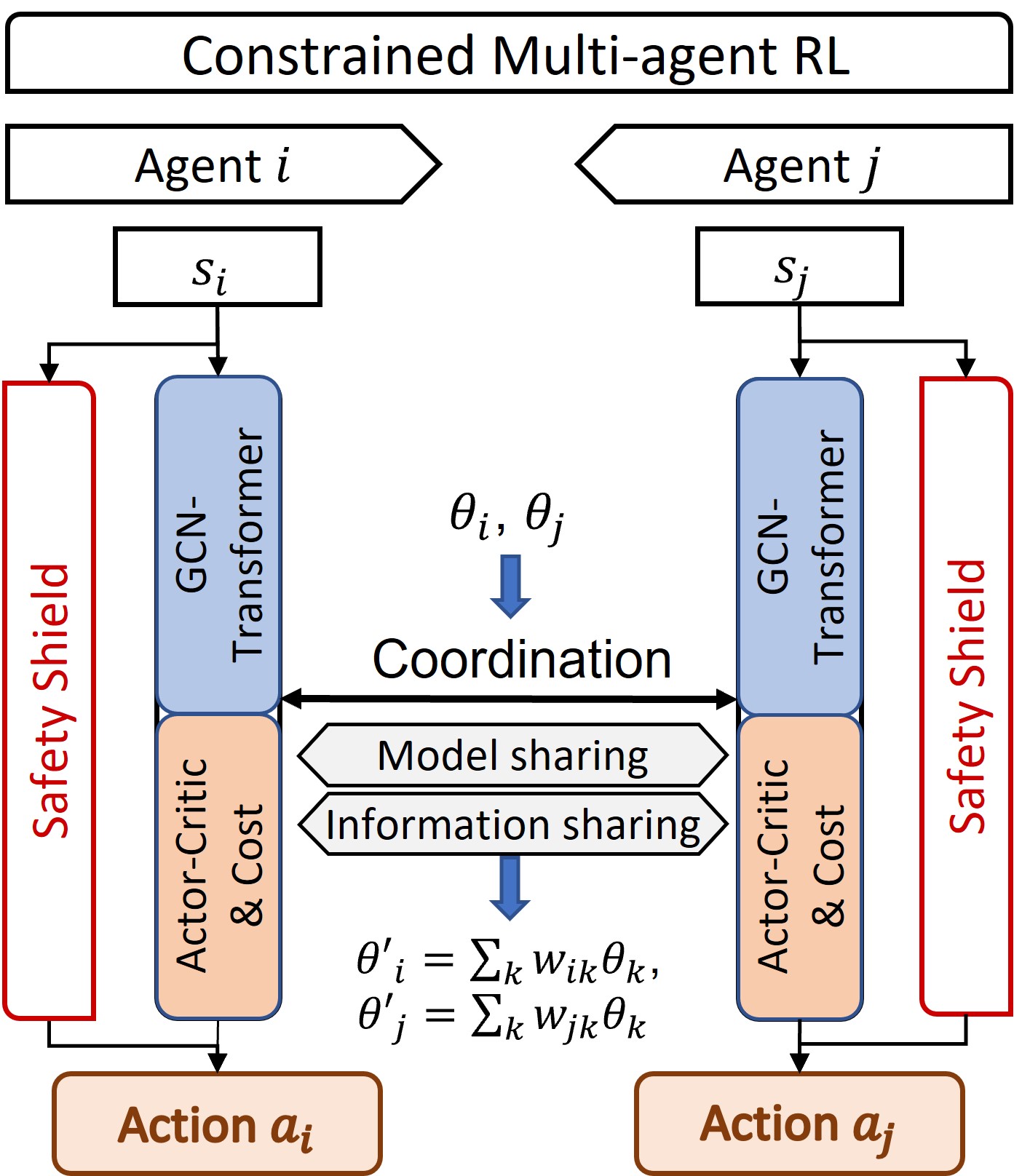}
\label{fig:framework}}
\label{fig:model}
\vspace*{-0.1cm}
\caption{Single agent's model pipeline in~\ref{fig:pipeline}; Constrained MARL framework in~\ref{fig:framework}}
\vspace*{-0.4cm}
\end{figure*}

In this section, we introduce our major contribution, the Constrained Multi-Agent Advantage Actor-Critic (C-MAA2C) algorithm with parallel CBF-based safety checking for collaborative policy learning of CAVs. The algorithm enhances the efficiency and safety of the system in challenging driving scenarios with the parallel usage of a CBF-based \textit{Safety Shield} module and constrained MARL as is shown in Fig.~\ref{fig:framework}. We will introduce the main algorithm C-MAA2C in subsection~\ref{sec:CMAA2C}, followed by details of the safety checking and training process in subsections~\ref{sec:Safety_Checking} and~\ref{sec:training}.
%Motivated by the framework of decentralized policy gradient for safe MARL~\cite{lu2021decentralized} and advantage actor-critic structure [A3C], 
\subsection{Constrained Multi-Agent Advantage Actor-Critic}
\label{sec:CMAA2C}
In Algorithm~\ref{alg:C-MAA2C}, we use centralized training decentralized execution design. Each agent maintains a policy network $\iactor{i}$ ("actor") with parameter $\boldtheta_i$, a $Q(s,a)$ network $\icritic{i}$ with parameter $\boldphi_i$ ("critic") for the reward $r_i(s,a)$ and another $Q^C(s,a)$ network $\icost{i}$ with parameter $\boldomega_i$ ("cost") for the cost $c_i(s,a)$ (as in Fig.~\ref{fig:pipeline}). $\boldtheta$ is defined as the parameter of the joint policy taken by all agents. The algorithm operates in forward view as agents interact within the environment. After observing the state $s_i$, the agent's stochastic policy computes for the probability over action set $\calP(\calA_i)$. Meantime, the safety checking based on $s_i$ generates the safe action set $\calA_i^{\safe}$ including all the safe candidate actions $a_{i,\kappa}$. The eventual behavior will be sampled from $\calP(\calA_i^{\safe})$ based on $\epsilon$-greedy. After all agents instruct their selected behavior $a_i$ to controller and have them executed, the algorithm synchronously goes to the next step by observing the reward $r_i (s,a)$, cost $c_i(s,a)$ and the new state $s'_i$. %Despite being private, the reward and cost signals $r_i$ and $c_i$ received by each agent are associated with the global state and action $s,a$. The CAV agents also want to maximize collective returns. 
Specifically, all the CAVs want to collaboratively optimize the total expected return of the system defined as $J^R(\boldtheta)=\frac{1}{n}\sum_{i\in\calN}\JR{i}{\boldtheta}$:
\iffalse
\begin{align}
    \JR{}{\boldtheta} &=\sum_{i\in\calN_i}\Expect_{a^{t+k}\sim\pi^{\boldtheta}(\cdot | s^{t+k})}{[\sum_{k = 0}^\infty(\gamma_r)^k r_i(s^{t+k},a^{t+k})]}  \label{Object_func}
\end{align}
\fi
\begin{align}
    \JR{}{\boldtheta} &=\sum_{i\in\calN_i}\Expect_{a^{k}\sim\pi^{\boldtheta}(\cdot | s^{k})}{[\sum_{k = 0}^\infty(\gamma_r)^k r_i(s^{k},a^{k})]}  \label{Object_func}
\end{align}

Maximizing objective~\eqref{Object_func} is equivalent to minimizing the negative of such value, subject to the cost constraint that each agent should satisfy, and the constrained MARL problem is defined as the following optimization problem 
%$\zeta_i$ locally applied to each agent as lower bounds for $\JC{i}{\boldtheta}$. $\JC{i}{\boldtheta}$ is the expected accumulated cost by agent $i$ provided that other agents use policy $\boldtheta_{-i}$.
\begin{align}
    \min_{\boldtheta} \quad& -J^R(\boldtheta) \label{min_max} \\
    s.t.\quad &J^C_i(\boldtheta) \geq \zeta_i, \forall i\in \calN;\quad\Theta_{i}=\Theta{j}, j\in\calN_i \nonumber
\end{align}
where $\Theta_{i}\coloneqq\boldtheta_i\times\boldtheta_{-i}$ is defined as the local copy of the policy $\boldtheta$ owned by agent $i$ according to~\cite{lu2021decentralized}. By the Lagrangian method~\cite{boyd2004convex}, the problem~\eqref{min_max} can be written as the following problem solved through the training process:
\begin{align}
    \min_{\boldtheta_i\in\boldtheta}\max_{\boldlambda\geq 0}&\quad \calL(\boldtheta_i,\boldtheta_{-i},\boldlambda_i,\boldlambda_{-i}) \\
    s.t.& \quad \Theta_{i}=\Theta{j} \quad, \forall j\in\calN_i, \forall i \nonumber
\end{align}
where $\boldlambda_i, \boldlambda_{-i}$ denote the dual variables, and $\calL(\boldtheta_i,\boldtheta_{-i},\boldlambda_i,\boldlambda_{-i}) \triangleq \frac{1}{n}\sum_{i\in\calN}[\JR{i}{\boldtheta}+ \dotprod{\zeta_i - \JC{i}{\boldtheta}}{\boldlambda_i}]$.
%\vspace{-5pt}
%\vspace{-0.4in}
%\vspace*{-0.6cm}

%\vspace{-10pt}
%\vspace{-0.2in}
%\vspace{-10pt}
%$g_i(\boldtheta_i)\triangleq \vec{c}_i-\JC{i}{\boldtheta_i}$ and 
\setlength{\textfloatsep}{0pt}% Remove \textfloatsep
\begin{algorithm}[!h]
\caption{Constrained Multi-Agent A2C} \label{alg:C-MAA2C}
% \small 
\SetAlgoLined
\textit{Initialize} replay memory $M = \union_i M_i$;
\textit{Initialize} actor, critic and cost networks $\boldtheta_i^0, \boldphi_i^0, \boldomega_i^0$; 
\textit{Initialize} $\boldvartheta_i^0=\mathbf{0}, \boldlambda_i^0 = 0$ \;
 \For {each episode $\epsilon$}{
    \textit{Initialize} $s=\prod_i s_{i}\in \calS$\;
    \textit{Initialize} safe action set $\calA^{\safe} =  \prod_i \calA_i^{\safe} = \calA$\;
    \For {each training cycle $\tau$}{
        \For{each step}{
            Choose $a_i \in \calA_i^{\safe}$ based on $\epsilon$-greedy, $a= \prod a_i$\;
            %Get the action $a = ASM(a)$ that pass the actuator's safety mapping $ASM$\;
            Execute action $a$, observe rewards $r=\{r_i\}$, costs $c=\{c_i\}$, and the new state $s'=\prod_i s'_{i}$ \; 
            Store $(s_i, a_i, r_i, c_i, s'_i),\forall i$ in $M_i$\;
            \lIf{collision}{continue}
            Update $\calA^{\safe} = \textbf{Safety\_Checking}(s')$\;
            $s \leftarrow s'$\;
        }
        Perform \textbf{Training}($\tau, M, \boldtheta_i^\tau, \boldphi_i^\tau, \boldomega_i^\tau, \boldvartheta_i^\tau, \boldlambda_i^\tau$)\;%in Alg.\ref{alg:Train}
    }
 }
\end{algorithm}

% this is the start of 4.B Safety checking
\subsection{Safety Shield and Safety Checking}
\label{sec:Safety_Checking}
To enhance the safety of agents during their interactions, we design a safety shield module to identify potential unsafe actions that violate the safety requirements and update the safe action set for the constrained MARL in Algorithm~\ref{alg:C-MAA2C}. Given agent $i$'s state $s_i$, safety checking will loop through all candidate actions $a_{i,\kappa}\in\calA_i$ and judge if $a_{i,\kappa}$ is safe based on control barrier functions and quadratic programming (CBF-QP). CBFs have been introduced to ensure set invariance with system dynamics knowledge~\cite{ames2016control,ames2019control} and ensure safe controller design of vehicles~\cite{he2021rule,wang2022ensuring, notomista2020enhancing,han2022behavior}. %In this work, We design CBFs based on our safety requirements and implement CBF  (CBF-QP) to suggest safe action candidates in advance of the decision period by MARL algorithm. 

Consider a nonlinear affine control system: $\dot{\vec{x}} = f(\vec{x})+g(\vec{x})\vec{u}$ with state $\vec{x} \in \mathbb{R}^n$, input $\vec{u} \in \mathcal{U} \subset \mathbb{R}^m$, $\mathcal{U}$ is the admissible input set of the system, $f$ and $g$ are locally Lipschitz. Define a superlevel set $ \mathcal{C} \subset \mathbb{R}^n$ of a differentiable function $h$: $\mathcal{C} = \{ \bm{x}\in \mathbb{R}^n : h( \bm{x}, t) \geq 0\}$. A set $\mathcal{C} \subset \mathbb{R}^n$ is \textit{forward invariant} if for every $\bm{x}_0 \in \mathcal{C} $, the solution $\bm{x}(t)$ to the system satisfies $\bm{x}(t) \in \mathcal{C} $ for all $t \geq 0$. The system is \textit{safe} with respect to the set $ \mathcal{C}$ if the set $\mathcal{C}$ is forward invariant \cite{ames2016control}. The function $h$ is a control barrier function (CBF) for the system on $\mathcal{C}$ if there exists $\gamma \in \mathcal{K}_{\infty}$ \cite{wu2016safety}: 
\begin{align}
\footnotesize
    \sup_{\vec{u} \in \mathcal{U}}[\frac{\partial h(\vec{x},t)}{\partial t} + L_f h(\bm{x},t)+L_g h(\bm{x},t)\bm{u}]\geq -\gamma h(\bm{x,t}) \nonumber
\end{align}

\begin{figure}[!t]
    \centering
    \vspace*{-8pt}
    \includegraphics[height=2.9cm]{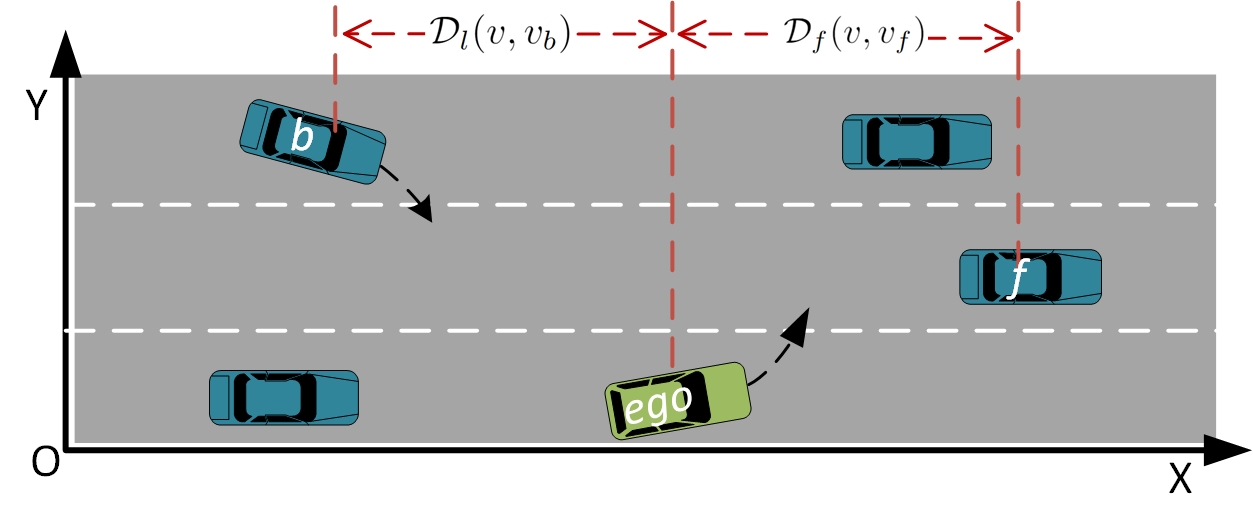}
    \vspace*{-14pt}
    \caption{Safety checking for ego vehicle's lane-change. We consider both the vehicle $f$ in the target lane and the vehicle $b$ entering the target lane.} %; $f,b$ are target vehicles
    \label{fig:safety}
    \vspace*{-8pt}
\end{figure}

CBF evaluating the safety of a candidate action $a_{i,\kappa}$ focuses on the relevant vehicles given $a_{i,\kappa}$ will be executed. As is shown in Fig.~\ref{fig:safety}, if a change-lane action is evaluated, target vehicles are the nearest neighbors from the front vehicles, front and rear vehicles on the target lane, and front and rear vehicles on the left/right other lane (if exist). Otherwise, only front and rear neighbors in the current lane are concerned. 
\setlength{\textfloatsep}{0pt}% Remove \textfloatsep
\begin{algorithm}[!t]
% \small 
\caption{Safety\_Checking} \label{alg:Safety_Checking}
\SetAlgoLined
Input: $s=\prod s_i$; initialize $\calA^{\safe}=\emptyset$\;
\For{each agent $i$}{
    \For{each action $a_{i,\kappa}\in \calA_i$}{
        \lIf{$a_{i,\kappa}$ is safe, \ie CBF-QP has a feasible solution} {append $a_{i,\kappa}$ to $\calA^{\safe}_i$}
    }
    \lIf{$\calA_i^{\safe}=\emptyset$}{$\calA_i^{\safe}=[Emergency\_stop]$}
}
\end{algorithm}

We adopt the widely used kinematic bicycle model for its simplicity while still considering the non-holonomic vehicle behaviors \cite{kong2015autonomous}. The state of the system $\vec{x}=[x, y, v, \psi]^T$ are the coordinates, velocity, orientation of the vehicle’s center of gravity (c.g.) in an inertial frame (X,Y). The inputs $\vec{u}$ to the system are acceleration at the vehicle’s c.g. $\alpha $ and the steering angle of the vehicle $\varphi$.

We consider the function of \textit{safety following distance} $\calD_f(v,v_f)$ defined as $\calD_f(v, v_f) = c_1 v + c_2(\frac{v^2}{2|\max(\alpha)|} - \frac{v_f^2}{2|\max(\alpha_f)|}) + D$, if the target vehicle is in the front (no matter which lane it is on), as in Fig.~\ref{fig:safety}. It takes the front and rear vehicles' velocity as input, and considers both the reaction delay term $c_1v$ and the hard-braking term $c_2(\frac{v^2}{2|\max(\alpha)|} - \frac{v_f^2}{2|\max(\alpha_f)|})$ which is proportional to the difference of hard-braking distances between the front and following vehicles, with an extra buffer distance as constant $D$. If target vehicle is behind, \textit{safety leading distance} is defined as $\calD_l(v, v_b) = c_1 v_b + c_2(\frac{v_b^2}{2|\max(\alpha_b)|} - \frac{v^2}{2|\max(\alpha)|}) + D$. Barrier function $h(\vec{x}, t)$ can then be respectively given as $h_f(\vec{x}, t) = (x_f-x) - \calD_f(v, v_f)$ and $h_b(\vec{x},t) = (x-x_b) - \calD_l(v, v_b)$.
For each candidate action $a_{i,\kappa}$, $\vec{u}_i$ is $a_{i,\kappa}$'s corresponding control input generated by a nominal controller, \eg PID controller. Then the safe candidate action can be evaluated by solving the below quadratic program \textit{CBF-QP}. If \textit{CBF-QP} is solvable, $a_{i,\kappa}$ is safe; otherwise it is unsafe.
\vspace*{-0.2cm}
\begin{align}
\textit{CBF-QP:}&\quad \min_{\vec{u} \in \mathbb{R}^m} \frac{1}{2}\parallel \vec{u}-\vec{u}_i \parallel^{2} \label{CBF_QP}\\
\textrm{s.t.} \quad & \frac{\partial h(\vec{x},t)}{\partial t} + L_f h(\bm{x},t)+L_g h(\bm{x},t)\bm{u}\geq -\gamma h(\bm{x}, t)\nonumber
\end{align}
%\vspace*{-0.2cm}

\vspace*{-10pt}
\subsection{Training}
\label{sec:training}
%\zhili{Here explain what the train algorithm is doing, basically following the safe-dec-PG; after explanation of the framework, followed by the details.}
As is preliminarily introduced in~\eqref{sec:CMAA2C}, the algorithm operates in forward view and updates the model parameters in every training cycle. During the training, the algorithm loops through agents and sequentially updates their policy parameters $\boldtheta_i$, the critic and cost network parameters $\boldphi_i,\boldomega_i$, the auxiliary policy gradient variables $\boldvartheta_i$ and the dual variable $\boldlambda_i$ with the training batch $B_{i}$ sampled from memory. Steps are given in algorithm~\ref{alg:Train}. 
% We will unfold the gradient computation details about actor-related parameters first, followed by the update rules of critic and cost networks.

\begin{algorithm}[!t]
\scriptsize 
%\small 
%\vspace*{-10pt}
\caption{Training} \label{alg:Train}
\SetAlgoLined
Input: $\tau, M, \boldtheta_i^\tau, \boldphi_i^\tau, \boldomega_i^\tau, \boldvartheta_i^\tau, \boldlambda_i^\tau$\;
\For{each agent $i$}{
    Update the $\boldtheta_i^{\tau+1}$ with ~\eqref{param_update}\;
    Sample a batch $B_{i}^\tau$ from $M$\;
    Update $\boldphi_i, \boldomega_i$ with~\eqref{cost_TD} respectively\;
    Calculate $\widehat{\grad}_{\boldtheta_i}f_i(\boldtheta_i^\tau, \boldlambda_i^\tau)$\;
    Update the $\boldvartheta^{\tau+1}_i$ by ~\eqref{gradient_update}\;
    Calculate $(\widehat{J}_i^C)(\boldtheta_i^{\tau+1})$\;
    Update the $\boldlambda^{\tau+1}_i$ by ~\eqref{dual_update}\;
}
\end{algorithm}

Let $F_i(\boldtheta, \boldlambda_i) \triangleq \JR{i}{\boldtheta}+\dotprod{\zeta_i-\JC{i}{\boldtheta}}{\boldlambda_i}\ , \forall i$. The estimated policy gradients regarding primal variables are
\begin{align}
    \hatgrad_{\boldtheta_i}F_i(\boldtheta, \boldlambda_i) =& \hatgrad_{\boldtheta_i}\JR{i}{\boldtheta} %+\beta\grad_{\boldtheta_i}H(\iactor{s_t})
    -\dotprod{\hatgrad_{\boldtheta_i}\JC{i}{\boldtheta}}{\boldlambda_i}, \forall i \label{grad_policy}
    \vspace*{-0.5cm}
\end{align}
and the policy gradients with respect to dual variables are
\begin{align}
    \hatgrad_{\boldlambda_i}F_i(\boldtheta_i, \boldlambda_i) = \zeta_i-\hat{J}_i^C(\boldtheta_i), \forall i
\end{align}

The update of actor's policy network follows the approach from safe-Dec policy gradient algorithm~\cite{lu2021decentralized}, in which every agent maintains a local policy with parameters $\boldtheta_i$ and a copy of auxiliary policy gradients $\boldvartheta_i$ computed based on local and neighbors' gradients. $\boldvartheta_i$ is initialized as $\boldvartheta_i^0=\vec{0}$. For each training round $\tau$, the new local policy $\boldtheta_i^{\tau+1}$ is updated with the current local policy $\boldtheta_i^\tau$, the local copy of policy gradients $\boldvartheta_i$ and policies shared by its neighbors $\{\boldtheta_j^\tau\}_{j\in\calN_i}$ as in~\eqref{param_update}. The update of policy gradient follows~\eqref{gradient_update}. In~\eqref{param_update},~\eqref{gradient_update}, $\sigma^\tau$ is the stepsize; $\mathbf{W}$ is weight matrix characterizing relations among nodes in $\calG$ introduced in~\cite{lu2021decentralized}.
\begin{align}
    %\theta_i^{r+1} &= \theta_i^r \nonumber \\
    \boldtheta_i^{\tau+1} =& \sum_{j\in \calN_i}\mathbf{W}_{i,j}\boldtheta_j^\tau - \sigma^\tau \boldvartheta_i^\tau
    \label{param_update}\\
    \boldvartheta_i^{\tau+1} =& \sum_{j\in \calN_i}\mathbf{W}_{i,j}\boldvartheta_j^\tau 
    + \widehat{\grad}_{\boldtheta_i}F_i(\boldtheta_i^{\tau+1}, \boldlambda_i^\tau) \nonumber \\
    &- \widehat{\grad}_{\boldtheta_i}F_i(\boldtheta_i^\tau, \boldlambda_i^\tau), \forall i
    \label{gradient_update}
\end{align}
We use the temporal difference error defined in~\eqref{cost_TD} for critic and cost respectively, to compute the loss for two networks. Specifically, $R_i-V_i(s)$, $R_i^{C}-V_i^C(s)$ are advantages \cite{pmlr-v48-mniha16} of the return and cost to compute gradients of policy network in $\hatgrad_{\boldtheta_i}\JR{i}{\boldtheta}$ and $\hatgrad_{\boldtheta_i}\JC{i}{\boldtheta}$ respectively in~\eqref{grad_policy}.
% previously defined in~\eqref{adv_R},~\eqref{adv_C}. 
\begin{equation}
    L_{i,critic} = (R_i^t - V_i(s^t))^2, L_{i,cost} = (R_i^{C,t} - V_i^C(s^t))^2 \label{cost_TD}.
\end{equation}
% \begin{align}
%     &L_{i,critic} = (R_i^t - V_i(s^t))^2\label{critic_TD} \\
%     &L_{i,cost} = (R_i^{C,t} - V_i^C(s^t))^2 \label{cost_TD}
% \end{align}
The update of dual variable $\boldlambda$ follows the approach in~\cite{lu2021decentralized} as~\eqref{dual_update}, where $\calP_\Lambda$ is the projection operator mapping $\boldlambda_i$ to a non-negative value and $\Lambda = \{\boldlambda_i | \boldlambda_i \geq 0\}, \forall i$ stands for the feasible set of $\boldlambda_i$; $\rho$ is the stepsize.
\begin{align}
    \boldlambda_i^{\tau+1} &= \calP_\Lambda ((1-\rho\gamma^\tau)\boldlambda_i^\tau + \rho\widehat{\grad}_{\boldlambda_i}F_i(\boldtheta_i^{\tau+1}, \boldlambda_i^\tau)
    \label{dual_update}
\end{align}

%\subsection{Pseudo-code}

\setlength{\textfloatsep}{0pt}

\section{Experiments and Evaluations}
\label{sec:experiment}
We deploy our experiment in the CARLA Simulator environment~\cite{Dosovitskiy17}, where each vehicle is configured with inborn GPS and IMU sensors and a collision sensor that detects the collision with other objects. We set the communication range of all CAVs as $100m$, within which another vehicle's information will be part of $o_i$ or shared observations $o_{j \in \calN_i}$ and $o_{\INF}$ and used by the ego vehicle in the decision-making and CBF safety-checking as introduced in Section~\ref{sec:prob}. The $k$-discretized throttle ranges in action space $\calA_i$ is universally set as $k=3$. We set the training cycle as every 16 steps, and set the discount factors $\gamma_r=0.99, \gamma_c=0.9$ accordingly. The constraints $\zeta_i$ for agents are universally $10$; the weight matrix $\mathbf{W}$ in training generally balances the weights between ego and others' policies while taking different values based on the number of agents. The training and testing of our algorithm and baselines took place in a server configured with AMD Ryzen 3970X 32-Core processor and four NVIDIA Quadro RTX 6000 GPUs. The experiments are performed with CARLA 0.9.11, Python 3.7, PyTorch 1.10, and CUDA 11.4.

\subsection{Simulation with Challenging Scenario}
We aim to deal with challenging scenarios in real life. Specifically, safety-critical events such as \textbf{\textit{running a red light}} at an intersection, 
 and \textbf{\textit{hard-braking}} in highway traffic incurred by another vehicle are usually immediate life threats to drivers and passengers. In the experiment, apart from the connected autonomous vehicles (\textit{CAVs}) and unconnected vehicles (\textit{UCVs}), we explicitly define a hazard vehicle (\textit{HAZV}) taking the aforementioned dangerous behaviors in 3 respective scenarios as illustrated in Fig.~\ref{fig:Scenarios} and~\ref{fig:highway_hard}. 

\subsubsection{Intersection} The figures in the first row of Fig.~\ref{fig:Scenarios} are predefined challenging intersection scenarios, where three CAVs (green) are driving through the intersection and the HAZV (red) from the crossing direction recklessly passes the intersection at the same time. The throttle values taken by the HAZV in simulator are randomly sampled from $[0.65,0.85]$ plus a tiny step-wise perturbation for continuous acceleration. In Fig.~\ref{fig:Scenarios} we present samples of initialization, success and failure cases of collision avoidance in the experiment.
\subsubsection{Highway} Figures in the second row of Fig.~\ref{fig:Scenarios} illustrate the challenging highway scenario, in which three CAVs, a UCV (red) and a HAZV (yellow mark) are spawned to ride on a multi-lane highway. The HAZV suddenly hard-brakes, taking the step-wise brake values in simulator randomly sampled from $[0.9,1.0]$ and causing an immediate threat to its rear CAVs. Meanwhile the UCV stays in its lane and takes throttles in $[0.3,0.7]$ securing its smooth driving. Success and failure of collision-avoidance cases are given in Fig.~\ref{fig:Scenarios}.
\subsubsection{Highway-Hard} 
\begin{figure}[!t]
\vspace*{-5pt}
\centering
\subfloat[]{\centering\label{fig:HwyH_init}\includegraphics[height=2.8cm,width=2.8cm]{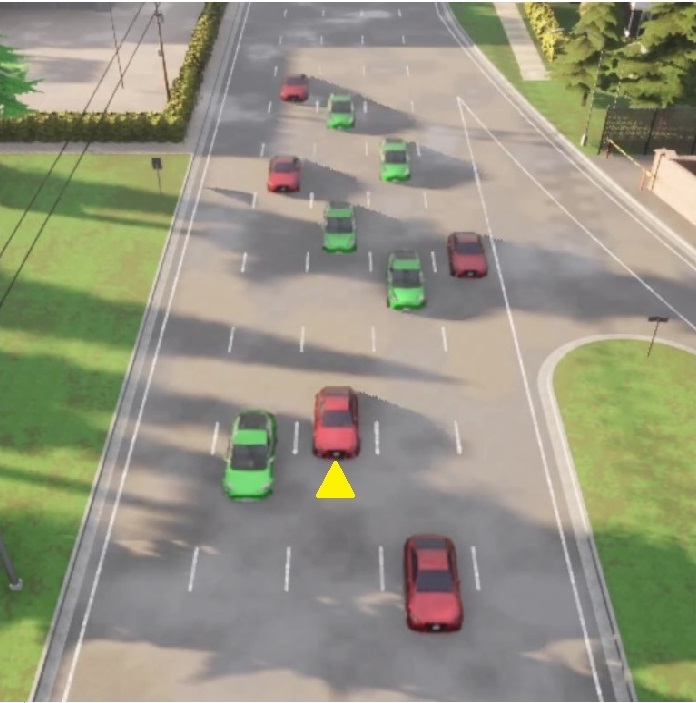}}\hfill
\subfloat[]{\centering\label{fig:HwyH_succ}\includegraphics[height=2.8cm,width=2.8cm]{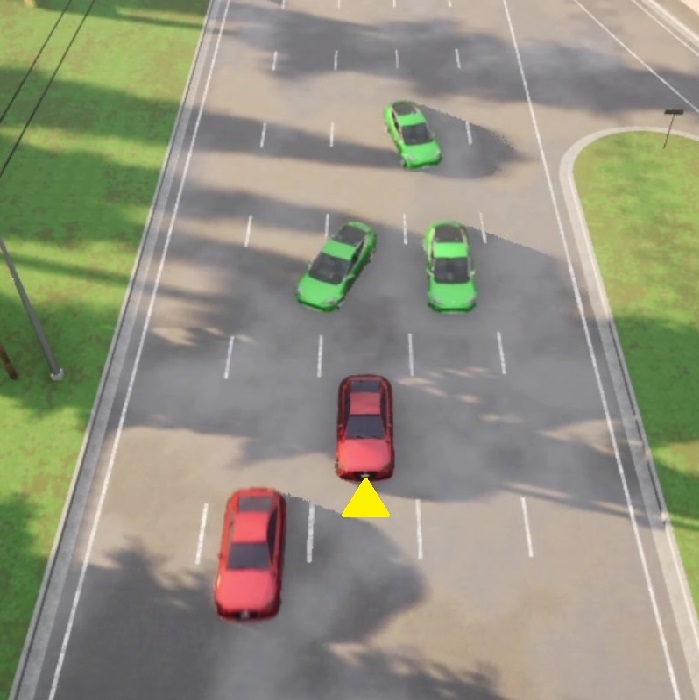}}\hfill
\subfloat[]{\centering\label{fig:HwyH_fail}\includegraphics[height=2.8cm,width=2.8cm]{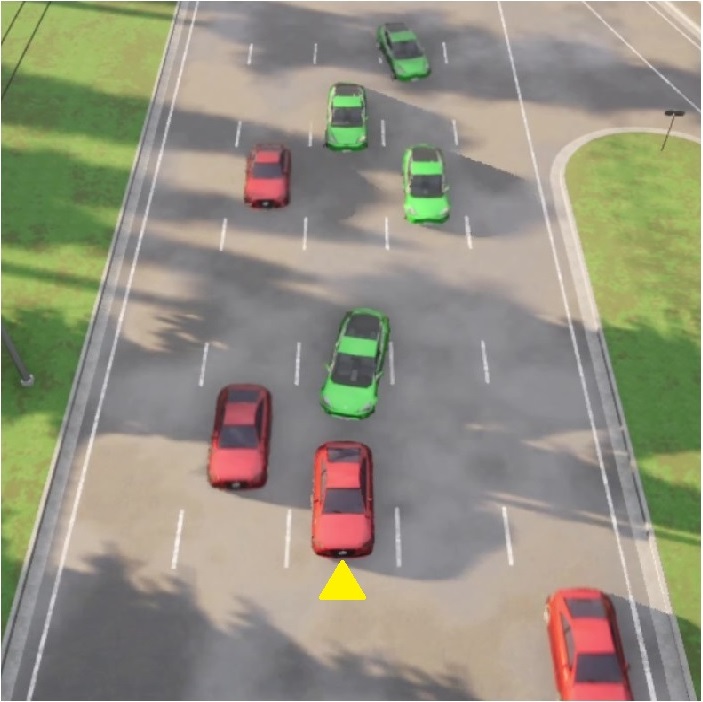}}
%\vspace*{-0.2cm}
\caption{\textit{Highway-Hard} scenario for testing, with 10 vehicles.~\ref{fig:HwyH_init}: \textit{Highway-Hard} initialization;~\ref{fig:HwyH_succ}: collision-free case with our method where agents collaboratively change to different neighboring lanes to avoid the hard-braking HAZV;~\ref{fig:HwyH_fail}: CAV agent in baseline collides with HAZV.}
\label{fig:highway_hard}
\vspace*{-10pt}
\end{figure}

For testing, we also devised a more difficult \textit{Highway-Hard} scenario shown in Fig.~\ref{fig:highway_hard}. Ten vehicles including 5 CAVs, 4 UCVs (red) and 1 HAZV (with yellow mark) are spawned in a compact traffic. The HAZV and UCVs behave similarly as in \textit{Highway}. The \textit{Highway-Hard} is comprehensively more challenging as it contains more agents and UCVs, and the compact vehicles' configuration produces complex interactions.
%\vspace*{-0.3cm}
\begin{table}[!t]
\centering
\begin{threeparttable}
\caption{Training Results in Two Scenario}
\begin{tabularx}{\columnwidth}{XXXc}
\hline
Scenario     & \multicolumn{2}{c}{Baselines}           & Ours        \\ \hline
             & w/o SS$^1$      & FC-CA2C$^2$     & GT-CA2C$^3$     \\ \hline
\textit{Intersection} & 21\%; 430.8 & 93\%; 572.8 & \textbf{96\%; 624.8} \\ \hline
\textit{Highway}      & 0\%; 166.4  & 91\%; 920.1 & \textbf{95\%; 955.6} \\ \hline
%\caption{Advantage of our method are highlighted.}
\end{tabularx}
\begin{tablenotes}
\item  $^1$w/o SS: without Safety Shield; $^2$FC-CA2C: Fully-Connected Constrained Advantage Actor-Critic; $^3$GT-CA2C: GCN-Transformer Constrained Advantage Actor-Critic
\item  Each entry above is (\textit{collision-free rate}; \textit{mean episode return}). Our method achieves \textbf{highest} safety and efficiency metrics in the training phase.
\end{tablenotes}
\label{tab:Train_results}
\end{threeparttable}
\vspace*{-0.4cm}
\end{table}

\begin{table}[!t]
\centering
\begin{threeparttable}
%\vspace*{-0.2cm}
\caption{Testing Results in Three Scenarios.}
%\vspace*{-0.2cm}
%\begin{tabularx}{m{0.27\linewidth}ccc}
\begin{tabularx}{\columnwidth}{m{0.27\linewidth}XXc}
\hline
Scenario     & \multicolumn{2}{c}{Baselines}   & Ours        \\ \hline
             & w/o SS      & FC-CA2C     & GT-CA2C     \\ \hline
\textit{Intersection} & 20\%; 444.8 & 86\%; 579.6 & \textbf{94\%; 586.8} \\ %\hline
\textit{Highway}      & 2\%; 185.3  & \textbf{90\%}; 922.6 & \textbf{90\%; 926.7} \\ %\hline
\textit{Highway-Hard}     & 0\%; 108.6  & 70\%; 706.4 & \textbf{78\%; 724.3} \\ \hline
\textit{Intersection} w/o Communication     & 20\%; 432.3  & 44\%; 473.7 & 44\%; 513.9 \\ %\hline
\textit{Highway-Hard} w/o Communication     & 0\%; 110.8  & 46\%; 567.5 & 48\%; 565.6 \\ \hline
\end{tabularx}
%\medskip
\begin{tablenotes}
\item  Each entry above is (\textit{collision-free rate}; \textit{mean episode return}). Our method \textbf{outperforms} baselines in two metrics, proving the improved safety and efficiency with GCN-Transformer and Safety Shield.
\end{tablenotes}
\label{tab:Test_results}
\end{threeparttable}
\end{table}

\subsection{Experiment Results}
 We trained our model (GCN-Transformer Constrained Advantage Actor-Critic; 'GT-CA2C' in the table~\ref{tab:Train_results},~\ref{tab:Test_results}), a baseline using our model without Safety Shield ('w/o SS' in tables) and another baseline 'FC-CA2C' with fully-connected layers (replacing GCN-Transformer), constrained advantage actor-critic and safety shield, each on \textit{Intersection} and \textit{Highway} scenarios. Our method and baselines are all under the multi-agent framework in Alg.~\ref{alg:C-MAA2C}. Training and testing experiment results are presented in table~\ref{tab:Train_results} and~\ref{tab:Test_results}. We \textbf{highlight} our method's top leading performance among all solutions. For each entry in tables, the left percentage is the collision-free rate in simulation; the right number is the mean episode return defined as the mean of agents' sums over stepwise rewards in every episode: $\sum_{\epsilon=1}^{m}{\avg{i}{\sum_{t}{r_i^t}}}/m$. Examples of episode return values from testing our model in \textit{Intersection} are given in the scatter plot in Fig.~\ref{fig:scatter}.

\subsubsection{Effectiveness of Safety Shield}
In all scenarios, our approach outperforms baselines in collision-free rate and overall return. Compared with the baseline 'w/o SS', the huge gaps in both metrics demonstrate improved safety and efficiency with our CBF-based safety checking method. 

\subsubsection{GCN-Transformer and Improved Environment Awareness}
 With the GCN-Transformer module applied compared to 'FC-CA2C', our method has leading performance in collision-free rates and mean episode returns in all three testing scenarios. In \textit{Highway-Hard} particularly, we find the advantage of our method is enlarged compared with the easier \textit{Highway}, and this verifies the significance of enhanced environment awareness with our approach under the more challenging and hazardous scenarios.

\subsubsection{Benefits of Coordination under Challenging Scenarios}
To verify the benefits of the coordination mechanisms, we test our model against the absence of V2X communication and observe the cascading performance without it in all solutions. In \textit{Intersection} scenario, the HAZV information becomes unavailable until it appears in CAVs' vision. In \textit{Highway-Hard} scenario, an ego vehicle is unaware of another CAV's intention to change lanes. From table~\ref{tab:Test_results} we could see, although our method surpasses the baselines in both metrics, the performance cannot match the excellence in test runs with coordinated communications. The collision-free rate drops from $94\%$ to $44\%$ in \textit{Intersection} scenario, and from $78\%$ to $48\%$ in \textit{Highway-Hard}, and this also applies to the baseline 'FC-CA2C'. The above results could prove the major contribution of coordination through information-sharing based on V2X communication.
%The consequence is two-fold. Without shared acceleration, agents have a worse estimate of safe distance $\calD$ in CBF; without shared lane-change, two agents could collide as they try to merge into the same lane from both sides. These could explain the noticeable drop of collision-free rate (from 78\% to 46\%, and from 70\% to 46\% in 'FC-CA2C') and returns as the absence of action-sharing in \textit{Highway-Hard} is more fatal to agents in a compact multi-lane scenario.
\iffalse
\begin{table}[!h]
\vspace*{0.2cm}
\caption{Testing Results in Normal Driving Scenarios}
\vspace*{-0.2cm}
\centering
\begin{tabular}{m{0.27\linewidth}ccc}
\hline
Scenario     & \multicolumn{2}{c}{Baselines}   & Ours        \\ \hline
             & w/o SC      & FC-CA2C     & GT-CA2C     \\ \hline
\textit{Intersection-Normal}     & 94\%  & 98\% & 100\% \\ %\hline
\textit{Highway-Normal}     & 96\%  & 98\% & 100\% \\ \hline
\end{tabular}
\label{tab:Test_results}
\end{table}
\fi

\subsubsection{Performance in Normal Driving Scenario}
\begin{figure}[!t]
\vspace*{-0.5cm}
\centering
\subfloat[]{\centering\label{fig:scatter}\includegraphics[width=4.2cm]{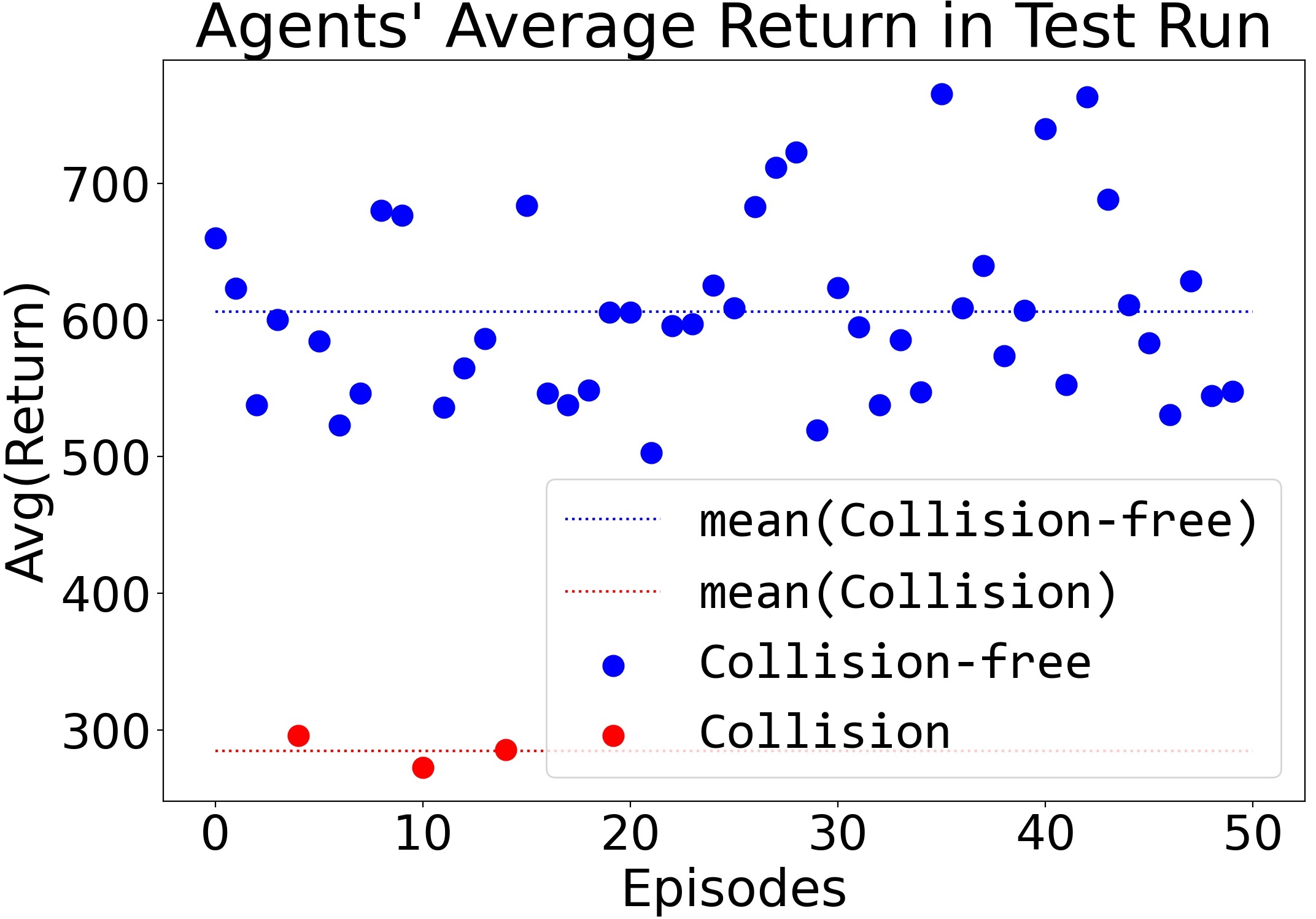}}\hfill
\subfloat[]{\centering\label{fig:bar}\includegraphics[width=4.3cm]{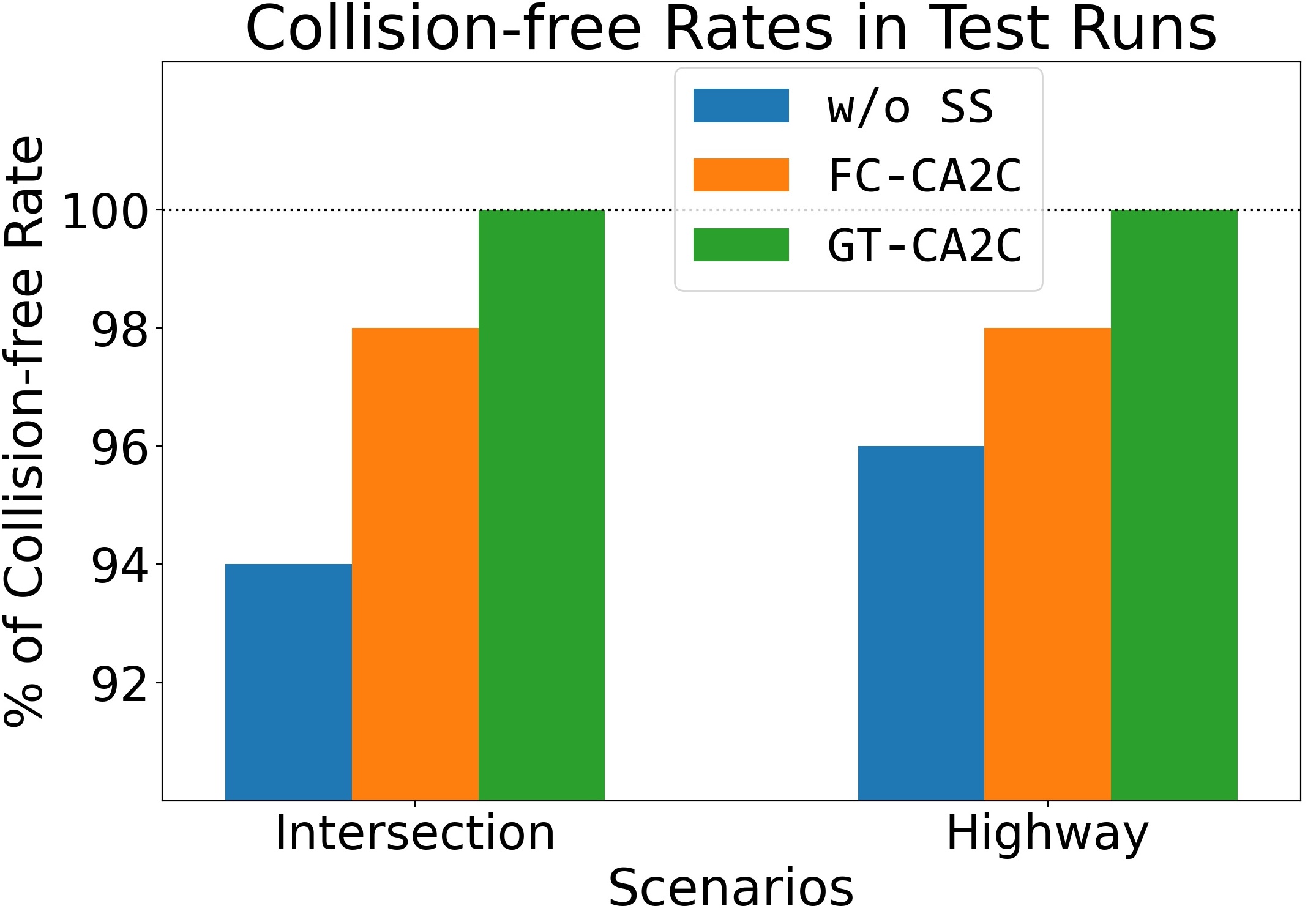}}
\vspace*{-0.2cm}
\caption{\ref{fig:scatter}: Scatter plots of episode returns in \textit{Intersection}; collision could affect agents' return greatly. \ref{fig:bar}: Collision-free rates in normal driving scenarios without hazard; our method achieves 100\% safety in both scenarios and leads all solutions.}
\label{fig:other_res}
\end{figure}

Lastly, we show results from testing in the remake hazard-free scenarios \textit{Intersection-Normal} and \textit{Highway-Normal} in Fig.~\ref{fig:bar}, in which the HAZV doesn't break into the intersection or brake abruptly. Our method can still perform well in the normal driving scenario as it achieved 100\% collision-free rate, while Baselines 'w/o SS' and 'FC-CA2C' both have collision without hazard.
%The scatter plots of episode returns from our algorithm's test runs in \textit{Intersection} and \textit{Highway} are given in Fig.~\ref{fig:scatter}, from which we could visualize the huge gap of returns between episodes with and without collisions.

%{coordination (shared information) -> coordinated decision (CMARL) \& coordinated safety checking (CBF); paragraph, coordination w/o coordination -> how they differ in reacting to the hazard, and the reward feedback; show either data / scenario, the reward/cost gap with successful and unsuccessful cases}
%\vspace*{-13pt}
\section{Conclusion}
\label{sec:conclusion}
In this work, we study the connected autonomous vehicles' cooperative policy-learning problem in challenging driving scenarios. We propose a constrained MARL coordinated policy learning framework with a safety shield for CAVs based on information-sharing. The GCN-Tranformer encoder is introduced to MARL to raise agents' spatial-temporal awareness of the environment. In experiments, we verify the effectiveness and advantage of our method and each of its modules in both safety and efficiency by comparing results with baseline models or settings, in challenging driving scenarios with hazard vehicles in traffic. Future work could extend to enhance the robustness of MARL algorithm and CBF safety shield with noisy and erroneous shared observations or models.

\iffalse
\begin{table}[!b]
\begin{tabular}{|c|c|ccc|c|}
\hline
Scenario     &                        & \multicolumn{3}{c|}{Baselines}                                                    & Ours        \\ \hline
             &                        & \multicolumn{1}{c|}{w/o SC}      & \multicolumn{1}{c|}{FC-A2C}      & FC-CA2C     & GT-CA2C     \\ \hline
Intersection & \multirow{2}{*}{Train} & \multicolumn{1}{c|}{21\%; 430.8} & \multicolumn{1}{c|}{90\%; 578.2} & 93\%; 572.8 & 96\%; 624.8 \\ \cline{1-1} \cline{3-6} 
Highway      &                        & \multicolumn{1}{c|}{0\%; 166.4}  & \multicolumn{1}{c|}{96\%; 948.2} & 91\%; 920.1 & 95\%; 955.6 \\ \hline
Intersection & \multirow{3}{*}{Test}  & \multicolumn{1}{c|}{20\%; 444.8} & \multicolumn{1}{c|}{90\%; 575.0} & 86\%; 579.6 & 94\%; 586.8 \\ \cline{1-1} \cline{3-6} 
Highway      &                        & \multicolumn{1}{c|}{2\%; 185.3}  & \multicolumn{1}{c|}{94\%; 920.7} & 90\%; 922.6 & 90\%; 926.7 \\ \cline{1-1} \cline{3-6} 
Highway*     &                        & \multicolumn{1}{c|}{0\%; 108.6}  & \multicolumn{1}{c|}{66\%; 574.7} & 70\%; 706.4 & 78\%; 724.3 \\ \hline
\end{tabular}
\end{table}
\fi

\bibliographystyle{IEEEtran}
{ \small 
\bibliography{sec_08_Ref}
}

\end{document}